\documentclass[a4paper]{article}
\usepackage[utf8]{inputenc} 
\usepackage[T1]{fontenc} 
\usepackage{lmodern} 
\usepackage{CJKutf8}
\usepackage{cite}
\usepackage[all]{xy}
\usepackage[dvips]{graphics,graphicx}
\usepackage{amsfonts,amssymb,amsmath,color,mathrsfs, amstext}
\usepackage{amsbsy, amsopn, amscd, amsxtra, amsthm,authblk}
\usepackage{enumerate,algorithmicx,algorithm,enumitem}
\usepackage{algpseudocode}
\usepackage[title]{appendix}
\usepackage{upref}
\usepackage{geometry,wrapfig}
\usepackage{amsthm,amsmath,amssymb}
\usepackage{relsize}
\usepackage{mathrsfs,mathtools}
\usepackage{bm,bbm}
\usepackage{ulem}
\usepackage{exscale}
\usepackage{tabularx}
\usepackage{threeparttable,multirow,dcolumn,booktabs}
\usepackage{algorithmicx,algorithm}
\usepackage{verbatim}
\usepackage{tabularx}
\usepackage{caption}
\usepackage{subfigure}
\usepackage[displaymath]{lineno}
\usepackage{float}
\usepackage{bm}
\usepackage{caption}
\usepackage{yhmath}
\usepackage{booktabs}
\usepackage{multirow}
\usepackage{makecell}
\usepackage[colorlinks,
            linkcolor=red,
            anchorcolor=red,
            citecolor=red
            ]{hyperref}
\usepackage{cleveref}

\usepackage[table,dvipsnames]{xcolor}

\definecolor{cgray}{gray}{0.80}
\newcolumntype{a}{>{\columncolor{cgray}}c}

\newcommand{\bs}[1]{\boldsymbol{#1}}
\newcommand{\bmu}{\bs{\mu}}

\newcommand{\bc}{\mathbf c}

\newtheorem{remark}{Remark}[section]

\numberwithin{equation}{section}
\usepackage{epstopdf}
\usepackage{pdfpages}

\begin{document}

\title{S$^2$GPT-PINNs: Sparse and Small models for PDEs}

\author{
Yajie Ji\footnote{School of Mathematical Sciences, Shanghai Jiao Tong University, Shanghai 200240, China. Email: {\tt jiyajie595@sjtu.edu.cn}. Y. Ji acknowledges the support from the NSFC (No. 124B2023). The code of S$^2$GPT-PINN is available at \url{https://github.com/DuktigYajie/S2GPT-PINN}.} ,\quad
Yanlai Chen\footnote{Department of Mathematics, University of Massachusetts Dartmouth, North Dartmouth, MA 02747. Email: {\tt{yanlai.chen@umassd.edu}}. Y. Chen is partially supported by National Science Foundation grant DMS-2208277 and by Air Force Office of Scientific Research grant FA9550-23-1-0037.}, \quad
Shawn Koohy \footnote{Department of Chemical and Biomolecular Engineering, University of Pennsylvania, Philadelphia, PA 19104, United States of America.}
}

\date{}
\maketitle

\begin{abstract}
We propose S$^2$GPT-PINN, a sparse and small model for solving parametric partial differential equations (PDEs). Similar to Small Language Models (SLMs), S$^2$GPT-PINN is tailored to domain-specific (families of) PDEs and characterized by its compact architecture and minimal computational power. Leveraging a small amount of extremely high quality data via a mathematically rigorous greedy algorithm that is enabled by the large full-order models, S$^2$GPT-PINN relies on orders of magnitude less parameters than PINNs to achieve extremely high efficiency via two levels of customizations. The first is knowledge distillation via task-specific activation functions that are transferred from Pre-Trained PINNs. The second is a judicious down-sampling when calculating the physics-informed loss of the network compressing the number of data sites by orders of magnitude to the size of the small model.
\end{abstract}


\section{Introduction}

In recent years, large-scale machine learning models have achieved numerous groundbreaking advancements, driving significant breakthroughs across various fields such as large language models (LLMs), image processing and many others \cite{vaswani2017attentionNew,brown2020language,he2016deep,deng2009imagenet, dhariwal2021diffusion}. Numerical methods for partial differential equations (PDEs) are no exception with the emergence of neural network based approaches \cite{HanJentzenE2018, raissi2019physics,devore2021neural}. However, the adoption of large models also comes with notable limitations exacerbated by the burgeoning parameter scale. Training such models requires vast amounts of data, computational resources, human effort, and often communication with the cloud, making them impractical in resource-constrained environments or when privacy is important. Furthermore, the high energy consumption of large models imposes significant economic burdens and raises concerns about environmental sustainability. These drawbacks highlight the importance of developing small models, which can run locally on a device and achieve comparable accuracy and robustness while significantly reducing computational costs and energy consumption. Recently, newly proposed small models have garnered significant attention \cite{schick2020s, lu2024blending,abdin2024phi,lu2024small}. By reducing storage and computational demands, small models address the challenges posed by large models and make machine learning  more practical and accessible for everyday applications essentially democratizing artificial intelligence.

Currently, neural network-based solvers for speeding up parametric problems can be broadly categorized into three approaches. The first category focuses on learning solution maps between two infinite-dimensional function spaces, also known as operator learning. By efficiently encoding inputs within a specific functional space, these methods enable rapid problem-solving. Representative examples include the Deep Operator Network (DeepONet) proposed by Lu et al. \cite{lu2021learning} and the Fourier Neural Operator (FNO) introduced by Li et al. \cite{li2021fourier}. These methods demonstrate strong performance in capturing complex solution maps. The second category is meta-learning, which seeks to optimize the initialization of neural networks for parametric PDEs, enabling models to adapt quickly to different parameter values and reducing the dependency on large training datasets. For instance, Narayan et al. \cite{penwardena2021physics} constructed a surrogate model that maps the parameters of a PDE to the parameters of Physics-informed Neural Networks (PINNs) using statistical and numerical techniques, outperforming generic meta-learning and interpolation approaches. Furthermore, Dong et al. \cite{huang2022meta} proposed a unsupervised mesh-free deep learning method called the Meta-Auto-Decoder, which implicitly encodes PDE parameters through latent vectors, allowing pretrained models to adapt efficiently to new PDE parameter values. The third category incorporates traditional dimensionality reduction techniques. For example, Hesthaven et al. \cite{hesthaven2018nonNNRBM} proposed POD-NN, which uses Proper Orthogonal Decomposition (POD) to reduce the dimensionality of the solution space and then employs a neural network to map the reduced parameter space to the solution space. Similarly, Manzoni et al. introduced DL-ROM and POD-DL-ROM \cite{fresca2021comprehensive,fresca2022pod} , which leverage autoencoders to directly learn the mapping from parameters to solutions. A common advantage shared by these three approaches is that, once the neural network is trained, the computational cost of prediction and solution is negligible. However, these methods often fail to fully leverage the intrinsic physical principles of the governing PDEs in the online stage. Moreover, they require large amounts of training data to improve the accuracy of surrogate model. This makes them less suitable for scenarios when embedding the underlying physics is key or where obtaining high-fidelity solution data is challenging.

To better embed prior physical information into neural network solvers in the parametric PDE (pPDE) setting, recent developments such as Generative Pre-Trained Physics-Informed Neural Networks (GPT-PINN) \cite{chen2024gpt} and its nonlinear model reduction version (TGPT-PINN) \cite{chen2024tgpt} have introduced a novel parametric system meta-learning framework based on the model reduction principles of the more traditional and mathematically rigorous reduced basis method (RBM) \cite{hesthaven2016certified, quarteroni_reduced_2016}. In the (T)GPT-PINN architecture, the outer network contains only a single hidden layer, where each hidden neuron’s activation function is a fully pre-trained neural network corresponding to the solution of a PDE for a carefully chosen parameter value during the offline stage. This design enables the network to inherit the efficiency and interpretability of RBM while leveraging the expressiveness of neural networks. Similar to RBM, (T)GPT-PINN employs a greedy algorithm to adaptively ``learn" the parameter dependencies of the system. Starting from scratch, the hidden layer is expanded one neuron at a time, with each neuron representing a new basis function corresponding to a pPDE solution. This iterative growth process ensures that the reduced neural network captures the essential features of the solution manifold. As a result, (T)GPT-PINN achieves accurate and efficient surrogate solutions for new parameter values during the online phase, demonstrating both precision and scalability in parametric system modeling. Notably, by introducing several key innovations, the newly proposed Entropy-Enhanced GPT-PINN \cite{ji2025egptpinn} efficiently captures complex parameter-dependent shock formations and interactions. EGPT-PINN accurately solves the viscosity solutions of some classic Burgers' and Euler equations using only a few neurons, without relying on any a priori knowledge of the equations or their initial conditions. This demonstrates the significant potential of generative pre-trained neural networks in solving complex pPDEs efficiently and effectively.

Based on this foundation, we propose the S$^2$GPT-PINN, a more resource-efficient and computationally superior algorithm, which incorporates a sparsification strategy into GPT-PINN dynamically selecting interpolation and residual minimization points. In addition, S$^2$GPT-PINN introduces an interpolation-residual-based initialization strategy. By utilizing interpolation information recorded during the construction of the reduced basis, the initial values of the neural network are set to facilitate faster and more stable convergence during training. This design further enhances the computational efficiency and robustness of the algorithm, making it highly effective for solving pPDEs. 
Although there have been adaptive collocation strategies \cite{lau2024pinnacle, MCCLENNY2023111722, gao2023failure} and approaches for changing the solution space \cite{ainsworth2021galerkin, siegel2023greedy} in the PINNs' framework, our proposed method differentiates itself by being, to the best of our knowledge, the first one capable of decreasing the solution space dimension to as small as its intrinsic dimension. Simultaneously, the number of collocation points is decreased to be on the same magnitude as the reduced solution space.

The remainder of this paper is organized as follows. In Section \ref{sec:bgd}, we review the RBM, PINN and GPT-PINN methods in the parametric setting. The main algorithm is given in Section \ref{sec:method} with numerical results  in Section \ref{sec:numerics}. We draw conclusions in Section \ref{sec:conclusion}.

\section{The Dense and Large Model: PINNs}
\label{sec:bgd}

We define our problem to be the following time-dependent parametric PDE on the spatial domain $\Omega \subset \mathbb{R}^d$ with boundary $\partial \Omega$, and parametric domain $ \mathcal{D} \ni \bmu$:
\begin{subequations}
\label{eq:pPDE}
\begin{equation}
\label{eq:pPDE-a}
\frac{\partial u(\bm{x}, t;\bmu)}{\partial t} +\mathcal{F}(u)(\bm{x}, t;\bmu)=0, \, \bm{x} \in \Omega, \, t \in[0, T],
\end{equation}
\begin{equation}
\label{eq:pPDE-b}
\mathcal{G}(u)(\bm{x}, t;\bmu)=0, \quad \bm{x} \in \partial \Omega, \quad t \in[0, T], 
\end{equation}
\begin{equation}
\label{eq:pPDE-c}
u(\bm{x}, 0;\bmu)=u_0(\bm{x};\bmu), \quad \bm{x} \in \Omega.
\end{equation}
\end{subequations}
Here $\mathcal{F}$ is a differential operator and $\mathcal{G}$ denotes a linear boundary operator. 

By embedding physical laws, typically expressed as PDEs, into the training process of deep neural networks (DNN), PINNs provide a flexible and efficient framework for solving a wide range of forward and inverse problems in science and engineering. DNNs consist of multiple layers of interconnected neurons, where each layer applies a sequence of linear transformations followed by nonlinear activation functions. The output of a DNN serves as an approximation to the PDE solution, denoted as $\Psi_{\mathsf{NN}}({\bm x}; \Theta)$ parameterized by the neural network parameter $\Theta$: 
\begin{equation}
\label{eq:DNN}
u(\bm x)\approx \Psi_{\mathsf{NN}}({\bm x}; \Theta) \coloneqq (C_L \circ \sigma \circ C_{L-1} \cdots \circ \sigma \circ C_1)(\bm x).
\end{equation}
where $\bm x$ is the input vector, $\Theta = \{\bm{W}_{\ell},\bm{b}_{\ell}\}_{{\ell}=1}^L$ represents the set of parameters with $\bm{W}_{\ell}$ and $\bm{b}_{\ell}$ respectively denoting the weight matrix and bias vector of the $\ell$-th ($1\leq \ell \leq L$) layer. The activation function $\sigma$ introduces nonlinearity, typically through functions such as ReLU, Tanh or other nonlinear mappings. Each linear transformation $C_{\ell}$ in the network is defined as:
$C_{\ell}(\bm x) = \bm{W}_{\ell} \bm x+\bm{b}_{\ell}$.

By incorporating the governing equation \eqref{eq:pPDE-a}, initial conditions \eqref{eq:pPDE-b}, and boundary conditions \eqref{eq:pPDE-c} into the loss function, PINNs leverage the expressiveness of DNNs to approximate solutions to PDEs.
For pPDEs like \eqref{eq:pPDE}, we express the network as 
 $\Psi_{\mathsf{NN}}^{\bmu}(\bm x,t; \Theta(\bmu))$ to emphasize its parametric and time dependence, and often abbreviate it as $\Psi_{\mathsf{NN}}^{\bmu}(\bm x,t)$ or simply $\Psi_{\mathsf{NN}}^{\bmu}$. 
\begin{equation}
\label{PDE-loss}
\begin{aligned}
\mathcal{L} & \left(\Psi_{\mathsf{NN}}^{\bmu}(\bm x,t);\bmu\right)=\left\|\frac{\partial}{\partial t} \Psi_{\mathsf{NN}}^{\bmu}+\mathcal{F}\left(\Psi_{\mathsf{NN}}^{\bmu}\right)\right\|_{L^2({\Omega \times (0,T]})}^2  \\
&+ \left\|\Psi_{\mathsf{NN}}^{\bmu}(0,t)-u_0\right\|_{L^2(\Omega)}^2 +\|\mathcal{G}\left(\Psi_{\mathsf{NN}}^{\bmu}\right)\|_{L^2({\partial \Omega \times [0,T]})}^2.
\end{aligned}
\end{equation}

\subsection{GPT-PINN}
GPT-PINN extends the concept of PINNs by introducing a parameter-efficient formulation inspired by the RBM. 

\subsubsection{Background: Reduced Basis Method}
In many real-world scenarios, there is a substantial demand for simulating systems associated with parameterized models in a multi-query or real-time fashion. These parameters may include system configurations, initial conditions, source terms, or level of uncertainties. Consider a parameterized PDE, $\mathcal{A}(u ; \bmu)=f, \, \text { in } \Omega$ 
where $\mathcal{A}$ is a parameterized operator together with necessary
boundary and initial conditions that depend on $\bmu \in \mathcal{D}$; 
$u \in X$ is the solution in a Banach space $X$; $f \in X^{\prime}$ is a source term in the dual space $X^{\prime}$.

Assume that we can use the full-order model (FOM) to seek the high-fidelity approximation $u_h(\cdot,\bmu) \in X_h$ in a finite-dimensional space $X_h$, by solving:
$$A_h(\bmu) u_h(\cdot,\bmu)=f_h,$$
where $A_h(\bmu) \in \mathbb{R}^{N_h \times N_h}$ and $u_h(\cdot,\bmu) \in \mathbb{R}^{N_h}$, with $N_h$ being the size of the discretized system. Evaluating 
$u_h(\bmu)$ for a large number of parameter queries can be computationally intractable due to the repeated invocation of the FOM. RBM seeks to alleviate this computational burden by constructing accurate and efficient surrogate models. It computes an approximation $u_N(\cdot,\bmu)$ from an $N$-dimensional subspace $X_N$ of $X_h$, such that $u_N(\bmu) \approx u_h(\bmu) \text { for all } \bmu \in \mathcal{D}$.

The combination of theoretical rigor \cite{BinevCohenDahmenDevorePetrovaWojtaszczyk} and computational efficiency makes RBM popular for multi-query systems.  Moreover, the offline-online decomposition framework is pivotal in achieving significant computational speedups.

\noindent{\textbf{Online Stage}}
-- In the online stage, a reduced basis space $X_N$ is used to rapidly compute approximate solutions for new parameter values $\bmu$. The solution is represented as $u_N(\bmu)=\sum_{i=1}^N c_i(\bmu) \zeta_i,$ 
where the coefficients $\left\{c_i(\bmu)\right\}_{i=1}^N$ are determined by solving a reduced-order system $A_N(\bmu) c_N=f_N,$ 
where $A_N(\bmu)=W^T A_h(\bmu) W$, $f_N=W^T f_h,$ 
and $W =\left[\zeta_1, \zeta_2, \ldots, \zeta_N\right] \in \mathbb{R}^{N_h \times N}$ is the basis matrix, obtained by applying the Gram-Schmidt (GS) orthogonalization process to the reduced basis space $X_N$. This enhances numerical stability and simplifies projection computations.

\noindent{\textbf{Offline Stage}} -- In contrast to other reduction techniques, such as POD, the offline stage of RBM requires a minimal number of full-order solutions, equal to the final dimension of the reduced basis space. The offline stage, briefly described in \Cref{alg:rbmoffline}, constructs a low-dimensional space $X_N$ by employing a greedy algorithm to select a set of representative parameters $\left\{\bmu_i\right\}_{i=1}^N$ and computing their corresponding high-fidelity solutions.

\begin{algorithm}[htbp]
\caption{A brief workflow for RBM offline stage}
\begin{algorithmic}[1]
\State Randomly select an initial parameter $\bmu^1$, and construct the initial reduced space: $X_1=\operatorname{span}\left\{u_h\left(\bmu^1\right)\right\}$.
\For {$i=2,3, \ldots, N$}
\State Solve for RB solution $u_{i-1}(\bmu)$ and calculate error estimator $\Delta_{i-1}(\bmu)$.
\State $\bmu^i=\operatorname{argmin}_{\bmu} \Delta_{i-1}(\bmu)$,
\State $X_{{i}}=\operatorname{span}\left\{u_h\left(\bmu^j\right), j=1,2, \ldots, i\right\}$.
\EndFor
\end{algorithmic} 
\label{alg:rbmoffline}
\end{algorithm}

\noindent{\textbf{Error Estimation}}
-- The efficient construction of the reduced basis space in the offline stage is primarily guided by an accurate and rigorous error estimator $\Delta_h(\bmu)$, which provides a rigorous bound for the error: $\left\|u_h(\bmu)-u_N(\bmu)\right\|_X \leq \Delta_N(\bmu)$.
The estimator is typically residual-based 
\begin{equation}
\label{eq:error_estimator}
\Delta_h(\bmu)=\frac{\left\|r_h\left(u_N(\bmu)\right)\right\|_{X^{\prime}}}{\beta(\bmu)},
\end{equation}
where $r_h(\bmu)=f_h-A_h(\bmu) u_N(\bmu)$ is the residual, $\beta(\bmu)$ is a stability constant dependent on the parameter. Error estimation serves as the cornerstone of RBM, guiding the offline construction of the reduced basis space and ensuring the reliability of online approximations. 

\subsubsection{The GPT-PINN}

PINNs' weaknesses include that training a vanilla PINN is usually significantly slower than a classic numerical method. To address this shortcoming, a meta-learning approach GPT-PINN was proposed in \cite{chen2024gpt}. Instead of training a separate neural network for each parameter 
$\bmu$, GPT-PINN leverages a pre-trained PINN as a general activation function within a reduced basis framework. In fact, GPT-PINN adopts a simplified ansatz, 
\begin{equation}
\label{eq:gpt-ansatz}
u_n(\bm{x},t)\approx\Psi_{\mathrm{\mathsf{NN}}}^{\Theta(\bmu)}(\bm{x}, t)=\sum_{i=1}^nc_i(\bmu)\Psi^{\bmu^i}_{\mathrm{\mathsf{NN}}}(\bm{x},t).
\end{equation}

This means that the GPT-PINN architecture, depicted in \Cref{fig:s2gptpinn} without the sparsification layer, is a network-of-networks, where the outer network is a single hidden layer with zero bias. Here, the width corresponds to the number of basis functions and $\{c_i(\bmu)\}_{i=1}^n$ acts as a weight vector. This structure is denoted by $\mathsf{NN}^s(2,{n},1)$ following \cite{chen2024gpt} with $2$ for $(x,t)$, ${n}$ for number of neurons in the hidden layer. The inner networks, serving as activation for the outer network, are full pre-trained PINNs instantiated by the PDE solutions at a  set of  adaptively-selected parameter values. These values are chosen by a mathematically reliable greedy algorithm from ground up.  That is, the small outer/meta layer adaptively grows its hidden layer one super neuron/network at a time. This adaptive process, dedicated to learning the parametric dependence of the system, represents the ``investment'' cost of GPT-PINN. With this initial investment, it is capable of providing significant computational savings in the multi-query and real-time settings thanks to the fact that their marginal cost is of orders of magnitude lower than that of an individual PINN.

The key feature, leading to that significant speedup, is the fact that the training cost of the meta network can be made independent of the size of the full PINN. The collocation sets $\mathcal{C}_o^r \subset \Omega \times(0, T), \mathcal{C}_{\partial}^r \subset \partial \Omega \times[0, T]$ and $\mathcal{C}_i^r \subset \Omega$ are used, similar to \Cref{PDE-loss}, to generate an approximation of the true loss. Finally, the terminal loss of the reduced network serves as the error indicator for GPT-PINN to expand its sole hidden layer from scratch via a RBM-like greedy algorithm.

\section{The Sparse and Small Model: S$^2$GPT-PINN}
\label{sec:method}

\subsection{Background: Reduced over-collocation}
The {RBM} is equipped with a rigorous a posteriori error estimator, typically constructed based on the residual of the original partial differential equation and the singular values of the differential operator \cite{quarteroni_reduced_2016}. However, direct computation of this estimator often relies on the high computational cost associated with the high-fidelity solution of the original problem via, for example, non-local numerical integrations. 

To address that, \cite{chen2021R2ROC} among others champion the more localized collocation approaches. The key idea is to construct a reduced model that approximates the solution manifold while leveraging carefully selected collocation points to ensure a balance between accuracy and computational cost. By enforcing the pPDE at the pre-selected collocation points $X^M(M\ll N_h)$, the reduced system is constructed as
$$
A_h^M\left(u_N(X^M ; \bmu)\right)=f(X^M),
$$
where $A_h^M \in \mathbb{R}^{M\times M}$ and $f^M = f(X^M)$.The approximate solution
 $u$ can be expressed in the following form based on the basis space $W$ and the collocation set $X^M$ {of cardinality $M$}:
$$
u_N^M = u_N(X_M ;\bmu)=\sum_{i=1}^N c_i(\bmu) \zeta_i(X^M)=\sum_{i=1}^N c_i(\bmu) \eta_i,
$$
where the coefficients $\left\{c_i(\bmu)\right\}_{i=1}^N$ are determined by solving a reduced over-collocation system:
$$
A_N^M(\bmu)\bm{c}_N^M=f_N^M,
$$
where $A_N^M = (W^M)^TA_h^M W^M \in \mathbb{R}^{N\times N}, A_h^M\in \mathbb{R}^{M\times M}$, $f_N^M =(W^M)^Tf^M, f^M = f(X^M)\in \mathbb{R}^N $, $W^M=\left[\zeta_1^M, \zeta_2^M, \ldots, \zeta_N^M\right] \in \mathbb{R}^{M \times N}=P_*W$,  where $P_*=[e_{i_1},e_{i_2},\cdots,e_{i_M}]\in \mathbb{R}^{M\times N_h}$, $e_{i_j}$ represents the unit vector corresponding to the $j$-th collocation point and $\zeta_i^M = \zeta_i(X^M),i = 1,\cdots,N$. 

As illustrated in Eq.~\eqref{eq:error_estimator}, the residual-based error estimator serves as the upper bound of the approximation error in classical RBM. Similarly for our setting, we consider the residual
$$
r_h^M(\bmu) = f_h-A_h(\bmu)\bm{c}_N^M.
$$

In \cite{chen2021R2ROC}, a novel error estimation based on reduced residuals was proposed, introducing the reduced residual reduced over-collocation method (R2ROC), which utilizes a hyper-reduced collocation strategy. This method leverages collocation points that are twice the number of basis functions, with half derived from basis function interpolation and the other half from the selected residuals of the corresponding partial differential equation. This innovative approach achieves simultaneous improvements in computational efficiency for both the offline and online stages. The R2ROC method is also divided into offline and online stages. During the offline stage, basis selection is guided by the proposed error estimator based on reduced residuals. The fact that the calculation of the reduced solution and the error estimator depends only on the points in the hyper{-reduced} collocation set leads to the effectiveness of the R2ROC algorithm. 

\subsection{Methodology}

\begin{figure}[htbp]
    \centering
    \includegraphics[width=0.9\linewidth]{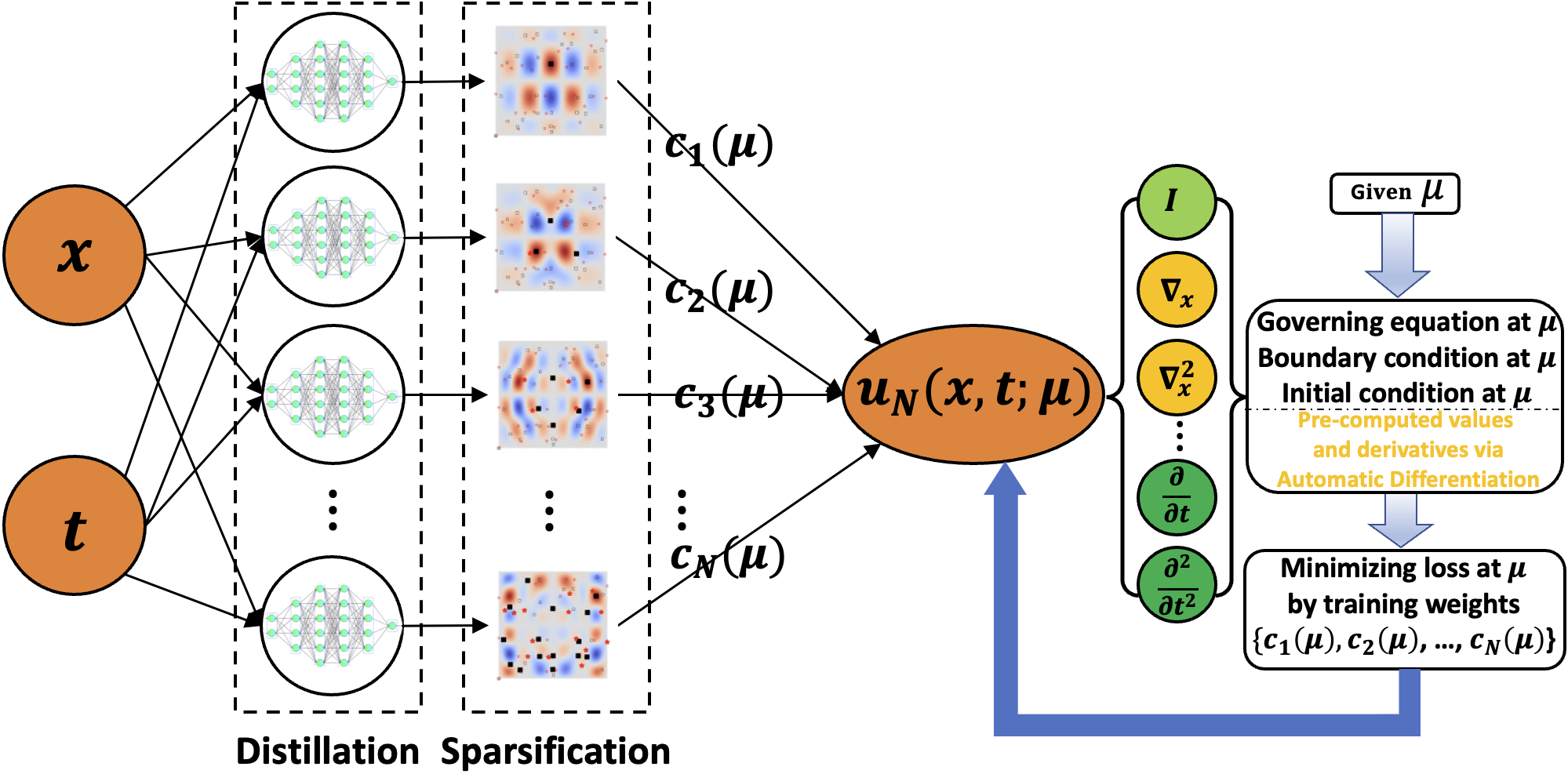}
    \caption{{Schematics of S$^2$GPT-PINN.}}
    \label{fig:s2gptpinn}
\end{figure}
Inspired by R2ROC and given the collocation nature of PINN and GPT-PINN, we propose in this work to further enhance GPT-PINN by a sparsification strategy and a novel reduced loss function detailed in \Cref{sec:online}. This loss function is based on the sparse collocation set and serves as an error indicator in the greedy algorithm to build the meta network. The effectiveness of the reduced loss function and the error indicator is wholly dependent on the choice of the over-collocation set denoted by $X^M$, so the key question is how to construct it. Interpolation-based points focus on regions of significant solution variation to capture the essential features of the solution manifold, while residual-based points target locations with the largest pPDE residuals to enforce accuracy in satisfying the governing equations. Together, these strategies, detailed in \Cref{sec:offline}, enable the construction of efficient and precise reduced models. 

The first subset of the hyper-collocation set, denoted as $X_s^N$, is derived from the points where the interpolation error of the basis functions in the reduced basis space is maximized. This ensures that the newly introduced hyper-reduced approximation is a generalized empirical interpolation approximation of the high-fidelity solution. 
The remaining part of $X^M$, denoted by $X_r^{N-1}$, is designed to control the residual of PDEs. It consists of points where the residuals of the basis functions' governing equations are largest. These points ensure that the reduced solution provides a good approximation satisfying the parametric partial differential equations across the parameter domain. Since the residual space spanned by the representative parameters 
$\bmu^n$ offers a stable interpolatory approximation for the residuals of any parameter $\bmu \in \mathcal{D}$. 

See \Cref{fig:s2gptpinn} for the S$^2$GPT-PINN schematic.

\subsubsection{Online stage - Knowledge Distillation}
\label{sec:online}

In the online stage, we assume that the representative parameters $\{\bmu^1, \bmu^2, \cdots, \bmu^{{n}}\}$, for some $n$ between 1 and the terminal S$^2$GPT-PINN width $N$, and their corresponding pre-trained PINNs $\{\Psi_{\mathsf{NN}}^{\bmu^1}, \Psi_{\mathsf{NN}}^{\bmu^2}, \cdots, \Psi_{\mathsf{NN}}^{\bmu^{{n}}}\}$ have already been obtained. The sparsification set $X^{m}=\{x_1^*,x_2^*,\cdots,x_M^*\} =X_s^{{n}} \cup X_r^{{n}-1} $ has already been determined during the offline stage, to be detailed in \Cref{sec:offline}. 

Once these are in place, we design the loss function specific to S$^2$GPT-PINN for a new parameter $\bmu$,  $\mathcal{L}_{\text{PINN}}^{\text{S$^2$GPT}}(X^m, \bc; \bmu)$ as folows. 
\[
\mathcal{L}_{\text{PINN}}^{\text{S$^2$GPT}}(X^m, \bc; \bmu) = \mathcal{L}_{\text{int}}(X^m, \bc;\bmu) +\mathcal{L}_{i}(\bc;\bmu),
\]
where 
\begin{align*}
\mathcal{L}_{\text{int}}(X^m, \bc;\bmu) & =\frac{\sum_{(\bm{x}, t) \in X^M}\left\|\frac{\partial}{\partial t}\left( \Psi_{\mathrm{\mathsf{NN}}}^{\Theta(\bmu)}\right)+\mathcal{F}\left( \Psi_{\mathrm{\mathsf{NN}}}^{\Theta(\bmu)}\right)\right\|_2^2}{\left|X^M\right|},\\
    \mathcal{L}_i(\bc; \bmu) & = \left\lVert\sum_{i=1}^N c_i(\bmu)(\sum_{\ell=1}^{i}\beta_{i\ell})-1\right\rVert_2^2.
\end{align*}
Here, $\bm{c}=(c_1(\bmu),c_2(\bmu),\cdots,c_n(\bmu))^T$ is the weights in the S$^2$GPT-PINN ansatz,
\[
u_n(X^m)\approx\Psi_{\mathrm{\mathsf{NN}}}^{\Theta(\bmu)}(X^m)=\sum_{i=1}^nc_i(\bmu)\Psi^{\bmu^i}_{\mathrm{\mathsf{NN}}}(X^m).
\]
Moreover, $\left( \beta_{i\ell}\right)$ is the transformation matrix when the pre-trained PINNs $\{\Psi_{\mathsf{NN}}^{\bmu^1}, \Psi_{\mathsf{NN}}^{\bmu^2}, \cdots, \Psi_{\mathsf{NN}}^{\bmu^{{n}}}\}$ are orthonormalized to be $\{\xi_i\}_{i=1}^n$. That is, we have $\xi_i = \sum_{\ell=1}^{i}\beta_{i\ell} \Psi_{\mathsf{NN}}^{\bmu^\ell}$.

\begin{remark}
The PDE residual part of the loss $\mathcal{L}_{\text{int}}(X^m, \bc;\bmu)$ is the standard PINN residual of the governing PDE restricted on $X^m$. The treatments of the initial and boundary conditions, reflected by the term $\mathcal{L}_{i}(\bc;\bmu)$, are similar. We focus on the former. We are penalizing the mismatch between the S$^2$GPT-PINN solution evaluated at $t=0$, $u_n(\bm{x},t=0;\bmu)$ and the given (and potentially $\bmu$-dependent) initial condition $u_0(\bm{x}; \bmu)$. However,
\[
{u}_n(\bm{x},0;\bmu) = \sum_{i=1}^n c_i(\bmu)\xi_i(\bm{x},0)
= \sum_{i=1}^N c_i(\bmu)\left(\sum_{\ell=1}^{i}\beta_{i\ell} \Psi_{\mathsf{NN}}^{\bmu^\ell}(\bm{x},0)\right)
\approx \sum_{i=1}^n c_i(\bmu)\left (\sum_{\ell=1}^{i}\beta_{i\ell} u_0(\bm{x}; \bmu^\ell) \right ).
\]
In the case $u_0(\bm{x};\bmu)$ is smooth with respect to $\bmu$, they can be effectively expanded by the Empirical Interpolation Method \cite{Barrault_Nguyen_Maday_Patera}. Toward that end, we assume that the EIM basis of the initial condition is $\{q_j(\bm{x})\}_{j=1}^J$ whose discrete version is denoted by the $J \times J$ matrix $\mathbb{Q}$. We further denote transformation (column) vector is $T^{\bmu}$. We then have that
\begin{align*}
\lVert u_n(\bm{x},0;\bmu) - u_0(\bm{x}; \bmu) \rVert & \approx \left \lVert \sum_{i=1}^n c_i(\bmu)\left (\sum_{\ell=1}^{i}\beta_{i\ell}  \mathbb{Q} T^{\bmu^\ell} \right) -\mathbb{Q} T^{\bmu} \right \rVert \\
& \le \lVert \mathbb{Q} \rVert \left \lVert \sum_{i=1}^n c_i(\bmu)\left (\sum_{\ell=1}^{i}\beta_{i\ell}   T^{\bmu^\ell} \right) - T^{\bmu} \right \rVert.
\end{align*}
When the initial condition is $\bmu$-independent, this difference can be simplified as 
$$\left \lVert\sum_{i=1}^N c_i(\bmu) (\sum_{\ell =1 }^i\beta_{i\ell})-1 \right\rVert_2.$$ 
We therefore simply take $\mathcal{L}_i = \left \lVert\sum_{i=1}^N c_i(\bmu) (\sum_{\ell =1 }^i\beta_{i\ell})-1 \right\rVert_2^2$ in the loss function.
\end{remark}

This loss is also adopted to evaluate the accuracy of the reduced and sparsified network $\mathsf{NN}^s(2,n,1)$ for approximating the PDE solution at $\bmu$,
\begin{equation}
\label{eq:sgpt-delta}
\Delta_{\text{S$^2$GPT}}^n(\bm{c}(\bmu)) \coloneqq \mathcal{L}_{\text{PINN}}^{\text{S$^2$GPT}}(\bm{c}(\bmu)) .
\end{equation}
For simplicity, we only emphasize the dependence on $\bc$ which is a function of $\bmu$. The online training of $\mathrm{\mathsf{NN}^s(2,n,1)}$, which is also invoked offline in the greedy algorithm gradually increasing $n$ from $1$ to $N$, is then simply 
\begin{equation}
\label{eq:update_c}
\bm{c}\longleftarrow{}\bm{c}-\delta_s\nabla_{\bm{c}}\mathcal{L}_{\text{PINN}}^{\text{S$^2$GPT}}.
\end{equation}
Here $\delta_s$ is the online learning rate. Therefore, the training and updates of the neural network $\mathsf{NN}^s(2,n,1)$ only rely on the information of 
$\mathcal{L}_{\text{PINN}}^{\text{S$^2$GPT}}$ at the sparsification set $X^m$.

\subsubsection{Offline stage}
\label{sec:offline}

Next, we present the offline stage in \Cref{alg:sgpt-offline}, where a greedy algorithm is employed to construct the reduced and sparsified neural network. 
\begin{algorithm}[H]
\caption{S$^2$GPT-PINN: Offline Distillation and Sparsification} 
\begin{algorithmic}[1]
\State Randomly selected $\bmu^1$ from $\Xi_{\text {train }}$, train a full PINN at $\bmu^1$ to obtain $\Psi^{\bmu^1}_{\mathrm{\mathsf{NN}}}$ and compute 
$\xi_1:=\Psi^{\bmu^1}_{\mathrm{\mathsf{NN}}}\left(X^{N_h}\right)$. \Comment{Initialization}
\State ($\xi_1$,$X_s^1, \sigma_1, i_1, \beta_1$) = GEIM(\{$\xi_1$\},[]). Set $P_*=\left[e_{i_1}\right]^T$. \Comment{Initial Sparsification}
\State Initialize $m=n=1, X^m=X_s^n, W_1=\left\{\xi_1\right\}, W_{1, m}=P_* W_1$ and $X_r^0=\emptyset$.
\For{$n=2, \ldots, N$}
\State For all $\bmu \in \Xi_{\text {train }}$, train $\mathrm{\mathsf{NN}}^s(2,n-1,1)$ and record $\Delta^{n-1}_{\text{S$^2$GPT}}(\bmu)$. \Comment{Parameter sweep}
\State Find ${\bmu^n=\operatorname{argmax}_{\bmu \in \Xi_{\operatorname{train}} \backslash\left\{\bmu^i, i=1, \cdots, n-1\right\}} \Delta^{n-1}_{\text{S$^2$GPT}}(\bmu)}$. \Comment{Greedy choice}
\State Train a full PINN at $\bmu^n$ to obtain 
$\Psi^{\bmu^n}_{\mathrm{\mathsf{NN}}}$, and $\xi_n:=\Psi^{\bmu^n}_{\mathrm{\mathsf{NN}}}\left(X^{N_h}\right)$. \Comment{FOM query}
\State Update the S$^2$GPT-PINN network by adding the new neuron to the hidden layer to construct $\mathrm{\mathsf{NN}}^s(2,n,1)$. \Comment{Network augmentation}
\State ($\xi_n$, $X_s^n, \sigma_n, i_1, \beta_n$) = GEIM($\{\xi_i\}_{i=1}^n$, $\{\sigma_i\}_{i=1}^{n-1}$). \Comment{Sparsification}
\State Form the full residual $r_{n-1}(\boldsymbol{x})=A_h\left(\bmu^n\right)\Psi_{\mathsf{NN}}^{\bmu^n}(\boldsymbol{x})-f\left(\boldsymbol{x}\right)$.
\State ($r_{n-1}$, $X_r^{n-1}, i_2$) = EIM($\{r_i\}_{i=1}^{n-1}, X_r^{n-2}$). \Comment{Residual sparsification}
\State Update $W_n=\left\{W_{n-1}, \xi_n\right\}, m=2 n-1, X^m=X_s^n \cup X_r^{n-1}, P_*=P_* \cup\left[e_{i_1}, e_{i_2}\right]^T$.
\EndFor
\end{algorithmic} 
\label{alg:sgpt-offline}
\end{algorithm}
The process begins by randomly selecting an initial parameter $\bmu^1$ from the training set $\Xi_{\text{train}}$ and solving the full PINN at $\bmu^1$ to obtain the corresponding high-fidelity solution $\Psi^{\bmu^1}_{\mathrm{\mathsf{NN}}}$. The first collocation point is determined by identifying the location of the maximum value of the solution $|\Psi^{\bmu^1}_{\mathrm{\mathsf{NN}}}|$, and the hyper-reduced neural network $\mathrm{\mathsf{NN}}^s(2,1,1)$ and the reduced basis space $W_1$ is initialized at this parameter.

\begin{algorithm}
\caption{Generalized EIM}
\begin{algorithmic}[1]
\Procedure{GEIM}{$\{\xi_i\}_{i=1}^n$, $\{\sigma_i\}_{i=1}^{n-1}$}\Comment{Input: $\{\xi_i\}_{i=1}^n$, functionals $\sigma_i$}
\State Find $\left\{\alpha_j\right\}$ such that ${\displaystyle \xi_n=\xi_n-\sum_{j=1}^{n-1} \alpha_j \xi_j}$ and $\{\sigma_i\left(\xi_n\right)=0\}_{i=1}^{n-1}$. \Comment{Orthogonalization}
\State  Find $\boldsymbol{x}_*^n=\operatorname{argmax}_{\bm{x}}\left|\xi_n(\boldsymbol{x})\right|$ and define $\sigma_n(\cdot)=\sigma_{\boldsymbol{x}_*^n}^{\bmu^n}(\cdot)$. \Comment{Localization}
\State Compute $\xi_n=\xi_n / \sigma_n\left(\xi_n\right)$. \Comment{Normalization}
\State $\beta_n = (1-\alpha \beta^T)/\sigma_n\left(\xi_n\right)$.
\State Update $X_s^n=X_s^{n-1} \cup\left\{\boldsymbol{x}_*^n\right\}$, and let $i_n$ be the index of $\boldsymbol{x}_*^n$. \Comment{Augmentation}
\State \textbf{return} $\xi_n$, $X_s^n$, $\sigma_n(\cdot)$, $i_n$, and $\beta_n$. \Comment{Return}
\EndProcedure
\end{algorithmic}
\label{alg:geim}
\end{algorithm}

Subsequently, the algorithm iterates to construct the reduced basis space and collocation point set. In each iteration, a reduced neural network $\mathrm{\mathsf{NN}}^s(2,n-1,1)$ is trained at all parameters in $\Xi_{\text{train}}$, and a new parameter $\bmu^n$ is selected based on the maximum of the error indicator $\Delta^{n-1}_{\text{S$^2$GPT}}(\bmu)$. 
\begin{equation}
\bmu^n=\operatorname{argmax}_{\bmu \in \Xi_{\operatorname{train}} \backslash\left\{\bmu^i, i=1, \cdots, n-1\right\}} \Delta^{n-1}_{\text{S$^2$GPT}}(\bmu).
\end{equation}
A full PINN is then trained at $\bmu^n$ to compute the next high-fidelity solution $\Psi^{\bmu^n}_{\mathrm{\mathsf{NN}}}$, which is orthogonalized, via the Generalized EIM process \cite{MadayMula2013, maday2015generalized} detailed in \Cref{alg:geim} against the current reduced basis. The algorithm then takes $\xi_n$, the orthonormalized version of the solution $\Psi^{\bmu^n}_{\mathrm{\mathsf{NN}}}$ with the orthonormalization performed under the functional $\{\sigma_i(\cdot)\}_{i=1}^{n-1}$, to augment the reduced space $W_n=\left\{W_{n-1}, \xi_n\right\}$. A new collocation point is added based on the maximum value of $\xi_n$. They correspond to the PDE operator instantiated at $\{\bmu^i\}_{i=1}^{n-1}$ \cite{chen2021R2ROC}.

Additionally, a residual-based collocation point is determined by analyzing the residuals of the PDE, ensuring that the reduced model accurately satisfies the governing equation {across the entire parameter domain}. These residuals are orthonormalized using the vanilla EIM \cite{Barrault_Nguyen_Maday_Patera} as detailed in \Cref{alg:eim}.
\begin{equation}
\label{eq:select_point2}
\boldsymbol{x}_{* *}^{n-1}= \operatorname{argmax}_{x \in X^{N_h} /\left\{X^m, x_n^n\right\}}\left|r_{n-1}(\boldsymbol{x})\right|.
\end{equation}
\begin{algorithm}
\caption{EIM}
\begin{algorithmic}[1]
\Procedure{EIM}{$\{\xi_i\}_{i=1}^n, X_r^{n-1}$}\Comment{Input: $\{\xi_i\}_{i=1}^n$ and existing interpol. pts}
\State Find $\left\{\alpha_j\right\}$ such that ${\displaystyle \xi_n=\xi_n-\sum_{j=1}^{n-1} \alpha_j \xi_j}$ and $\xi_n(X_r^{n-1})=0$. \Comment{Orthogonalization}
\State  Find $\boldsymbol{x}_*^n=\operatorname{argmax}_{\bm{x}}\left|\xi_n(\boldsymbol{x})\right|$. \Comment{Localization}
\State Compute $\xi_n=\xi_n / \xi_n(X_*^n)$. \Comment{Normalization}
\State Update $X_r^n=X_r^{n-1} \cup\left\{\boldsymbol{x}_*^n\right\}$, and let $i_2$ be the index of $\boldsymbol{x}_*^n$. \Comment{Augmentation}
\State \textbf{return} $\xi_n$, $X_r^n$, $i_2$. \Comment{Return}
\EndProcedure
\end{algorithmic}
\label{alg:eim}
\end{algorithm}

Both types of collocation points, snapshot-based and residual-based, are combined into the collocation set 
$X^{2n-1}=X_s^n \cup X_r^{n-1} $, dynamically expanding the collocation points and sparsifying the network at each iteration. The reduced neural network is updated iteratively by adding neurons corresponding to the new basis functions. This construction ensures that the reduced model achieves high accuracy with minimal computational cost, effectively balancing the training complexity and model precision. The {skeleton of the} algorithm is provided in \Cref{alg:sgpt-offline}.

\subsubsection{The initial guess}
The initialization of parameters plays a critical role in network training, even for hyper-reduced networks such as S$^2$GPT-PINN, which involve only a limited number of training parameters. Specifically for S$^2$GPT-PINN, we observe that during the construction of the reduced basis space, the newly added basis functions correspond to the interpolation residuals of the solution vector for the new parameter on the previously constructed basis space. As a result, the weights of different bases are not uniform and show a relationship with the interpolation coefficients. Therefore, during the generation of S$^2$GPT-PINN, it is crucial to record the interpolation residual 
$\bm{\alpha}$ of each new $n$-th basis in the previously constructed $(n-1)$-dimensional basis space. By setting $[\bm{\alpha},0]$ as the initial values for all S$^2$GPT-PINNs with a $(2,n,1)$ architecture, the algorithm achieves faster and more stable convergence. This initialization effectively leverages the underlying interpolation structure, enhancing both accuracy and efficiency.

\section{Numerical results}
\label{sec:numerics}

In this section, we present numerical results of the S$^2$GPT-PINN applied to {four} families of equations, the Klein-Gordon equation, the Allen-Cahn equation, the Burgers’ equation, and the Helmhotlz equation. The code for all these examples are published on GitHub at \url{https://github.com/DuktigYajie/S2GPT-PINN}.

The Klein-Gordon equation on $(x,t)\in[-1,1]\times[0,5]$ is parameterized by $(\alpha,\beta,\gamma)\in[-2,-1]\times[0,1]\times[0,1]$.
\begin{equation*}
\begin{aligned}
    u_{tt} + \alpha u_{xx} + \beta u + \gamma u^2 + x\cos{(t)} &- x^2\cos^2{(t)} = 0,\\ 
     u(-1,t) = -\cos{(t)}, \quad u(1,t)&=\cos{(t)},\\
    u(x,0)  &= x, \\
    u_t(x,0)  &=0.
\end{aligned}
\end{equation*}
The Allen-Cahn equation on $(x,t)\in[-1,1]\times[0,1]$ is parameterized by $(\lambda,\epsilon)\in[10^{-4},10^{-3}]\times[1,5]$.
\begin{equation*}
\begin{aligned}
    u_t - \lambda u_{xx} &+ \epsilon (u^3-u)  =0,\\ 
    u(-1,t)&=u(1,t) \\
    u_x(-1,t)&=u_x(1,t) \\
    u(x,0)&=x^2\cos{(\pi x)}.
\end{aligned}
\end{equation*}
The Burgers' equation on $(x, t) \in[-1,1] \times[0,1]$ is parameterized with one parameter, the viscosity $\nu \in[0.005,1]$.
\begin{equation*}
\begin{aligned}
u_t+u u_x-\nu u_{x x} & =0, \\
u(-1, t)=u(1, t) & =0, \\
u(x, 0) =-\sin &(\pi x) .\end{aligned}
\end{equation*}
Finally, the two-dimensional Helmholtz equation on $(x,y) \in [-1,1]\times [-1,1]$, 
\[
\Delta u(x, y)+k^2 u(x, y)=q(x, y),
\]
with a homogeneous Dirichlet boundary and $k$ being a constant, has the parameter-dependent solution as follows:
\begin{equation*}
u(x, y;a_1,a_2)=(x^2-1)(y^2-1)\sin \left(a_1 \pi x\right) \sin \left(a_2 \pi y\right)
\end{equation*}
where $(a_1, a_2)\in[1,2]\times[1,4]$.

\subsection{The setup}

To better illustrate the advantages of our algorithm, we conduct a comparative analysis with the GPT-PINN method. All network parameters for the full PINN for the first three examples were configured to match those used in GPT-PINN \cite{chen2024gpt}. The full PINN also remains consistent with the settings in GPT-PINN. However, we adopt L-BFGS which is a variation of the BFGS algorithm, a widely used quasi-Newton method. In our simulations, we set the learning rate to be 0.1 and found that only 50 epochs were sufficient to achieve convergence. Unlike other optimizer such as SGD and Adam, L-BFGS determines termination conditions for each epoch based on changes in the second-order derivatives. As a result, even with the same number of epochs, the actual training time for the model may differ. From the CPU testing times, it can be observed that the time required for different parameters exhibits slight variations, causing the cumulative runtime curve reported in the next section to deviate from being strictly linear.

\begin{table}[htbp]
\centering
\begin{tabular}{@{}lcc@{}}
\toprule
\textbf{} & \textbf{Collocation sizes} & \textbf{Network sizes} \\ \midrule
\textbf{PINN} & \hspace{-0.5cm}$(11024,20612,10200, 66436)$  & $(1801,50049,1341, 1341)$ \\ 
\textbf{GPT-PINN} &\hspace{-0.5cm} $(11024,20612,10200, 66436)$& {\bf Small}: $(12, 12, 10, 24)$ \\ 
\textbf{S$^2$GPT-PINN} & {\bf Sparse}: $(23, 23, 19, 47)$ & {\bf Small}: $(12, 12, 10, 24)$\\ \bottomrule
\end{tabular}
\caption{GPT-PINN is smaller than PINN, and S$^2$GPT-PINN is smaller and sparser than PINN. Four numbers in each parenthesis correspond to the four examples.}
\label{tab:comparison}
\end{table}
To highlight that S$^2$GPT-PINN is both small and sparse, we display 
the numbers of collocation points and network sizes for PINN, GPT-PINN, and S$^2$GPT-PINN for all {four} examples in \Cref{tab:comparison}. We use the RAM-AW approach of \cite{hou2023enhancing} as the full-order model for the fourth example. 
While the numbers of collocation points for the first three examples are very similar, this example is more complicated, resulting in a collocation set about 6 times larger. 
The speedup factors (\Cref{tab:H_time}), of GPT-PINN and S$^2$GPT-PINN over PINN, are in turn more dramatic for this example.

\begin{table}[!htbp]
\centering
\begin{tabular}{@{}lcccc@{}}
\toprule
 & \textbf{KG} & \textbf{AC}&\textbf{B}&\textbf{H} \\ \midrule
\textbf{PINN} &1&1&1&1 \\ 
\textbf{GPT-PINN} &7e-3 &7e-3&4e-3&8e-3 \\ 
\textbf{S$^2$GPT-PINN} &3e-3&3e-3&1e-3&2e-3\\ \bottomrule
\end{tabular}
\caption{{Online compute time, normalized by that of the corresponding PINN solver, for the four examples.}}
\label{tab:H_time}
\end{table}

\subsection{Results}

\begin{figure}[htbp]
\centering
\includegraphics[width=\linewidth]{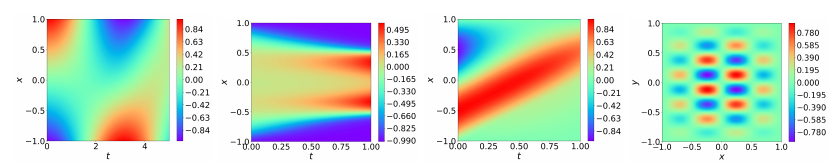}
\caption{The S$^2$GPT-PINN solution for the  four examples with $(\alpha,\beta,\gamma) =(-1.22,0.22,0.56) $, $(\lambda,\epsilon) = (0.00091,2.6)$, $\nu = 0.8212$, $(a_1,a_2)=(1.95,3.95)$.}
    \label{fig:s2gpt-solns}
\end{figure}

Here, we present numerical results for the {four} equations. When one figure has {four sub-}plots, we always follow the pattern that from top to bottom and/or left to right are Klein-Gordon, Allen-Cahn, Burgers', {and Helmholtz}. We first present, in \Cref{fig:s2gpt-solns}, sample S$^2$GPT-PINN solutions for one unseen parameter value per example.

\begin{figure}[thbp]
    \centering
\includegraphics[width=\linewidth]{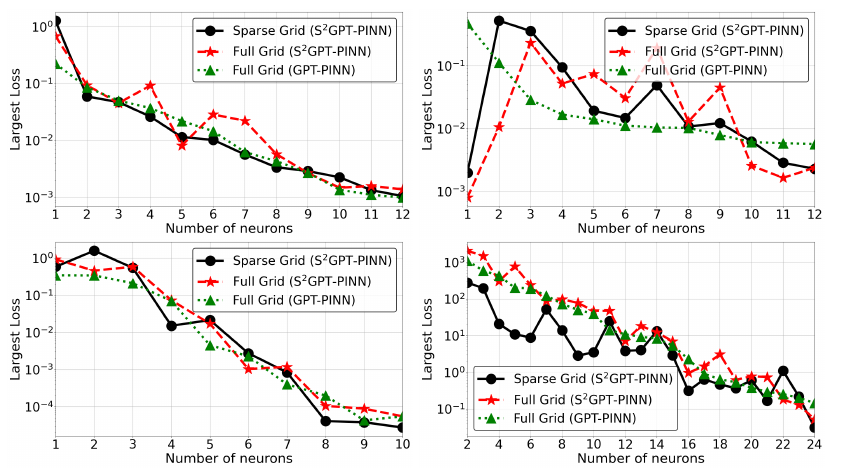}
    \caption{S$^2$GPT-PINN is as accurate as GPT-PINN: Worst case training losses for the S$^2$GPT-PINN (evaluated on the sparse and full grid) and GPT-PINN.}
    \label{fig:kgacb-loss}
\end{figure}


\vspace{-0.1cm}
Shown in \Cref{fig:kgacb-loss} are the history of convergence for the worst-case training loss, in comparison to that of the GPT-PINN, as we increase the size of the S$^2$GPT-PINN. Exponential convergence is noticeable. To verify that the sparsification still captures the loss accurately, we also record and display the loss on the full grid. We see that it corroborates the one on the sparse grid attesting to the effectiveness of our sparsification strategy.

\begin{figure}[thbp]
    \centering
\includegraphics[width=\linewidth]{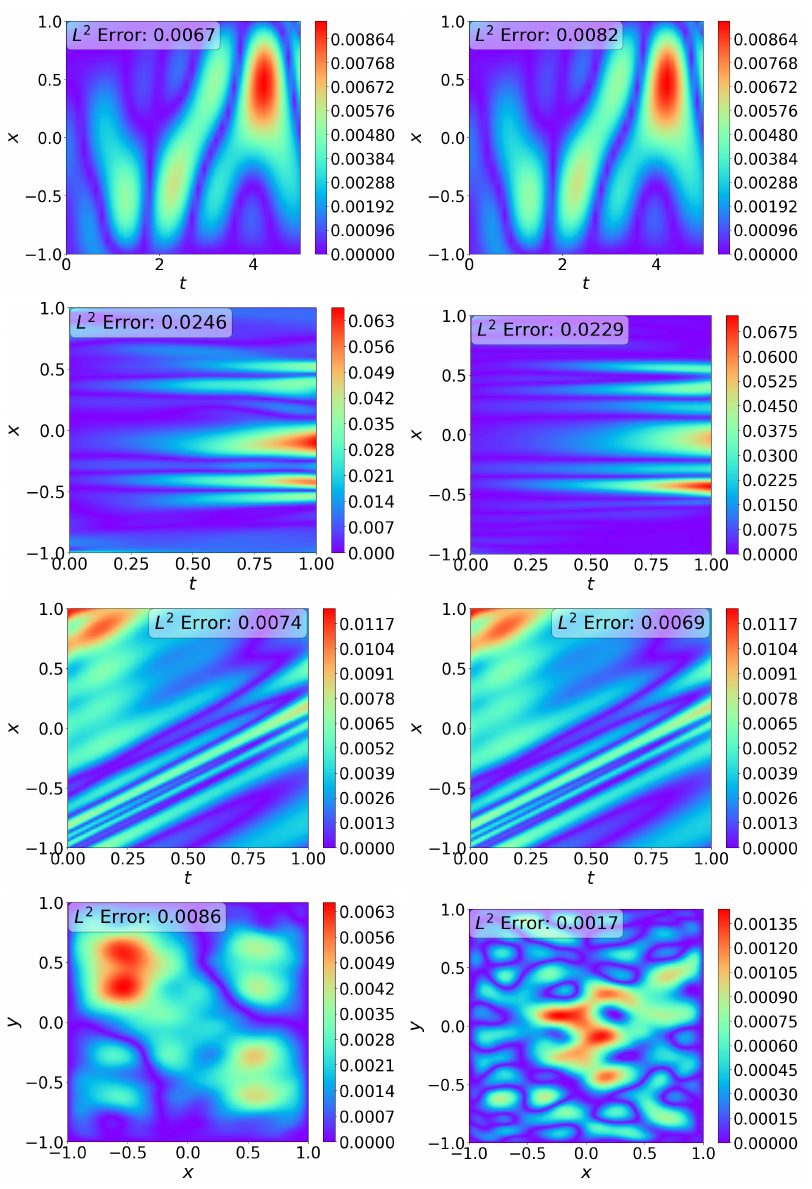}
    \caption{ S$^2$GPT-PINN (left) and GPT-PINN (right) pointwise errors and L2 error for a randomly selected test case. Parameter values for the four examples are $(\alpha,\beta,\gamma) =(-1.22,0.22,0.56) $, $(\lambda,\epsilon) = (0.00091,2.6)$, $\nu = 0.8212$, $(a_1,a_2)=(1.95,3.95)$.} 
    \label{fig:kgacBH-err}
\end{figure}

To further demonstrate S$^2$GPT-PINN's accuracy, we show in \Cref{fig:kgacBH-err} the actual pointwise errors at a random parameter value for all examples. It is clear that both the magnitude and the pattern are very similar between S$^2$GPT-PINN and GPT-PINN. This highlights the effectivity of S$^2$GPT-PINN since, as \Cref{tab:comparison} shows, S$^2$GPT-PINN is orders of magnitude sparser.

\begin{figure}[thbp]
    \centering
\includegraphics[width=\linewidth]{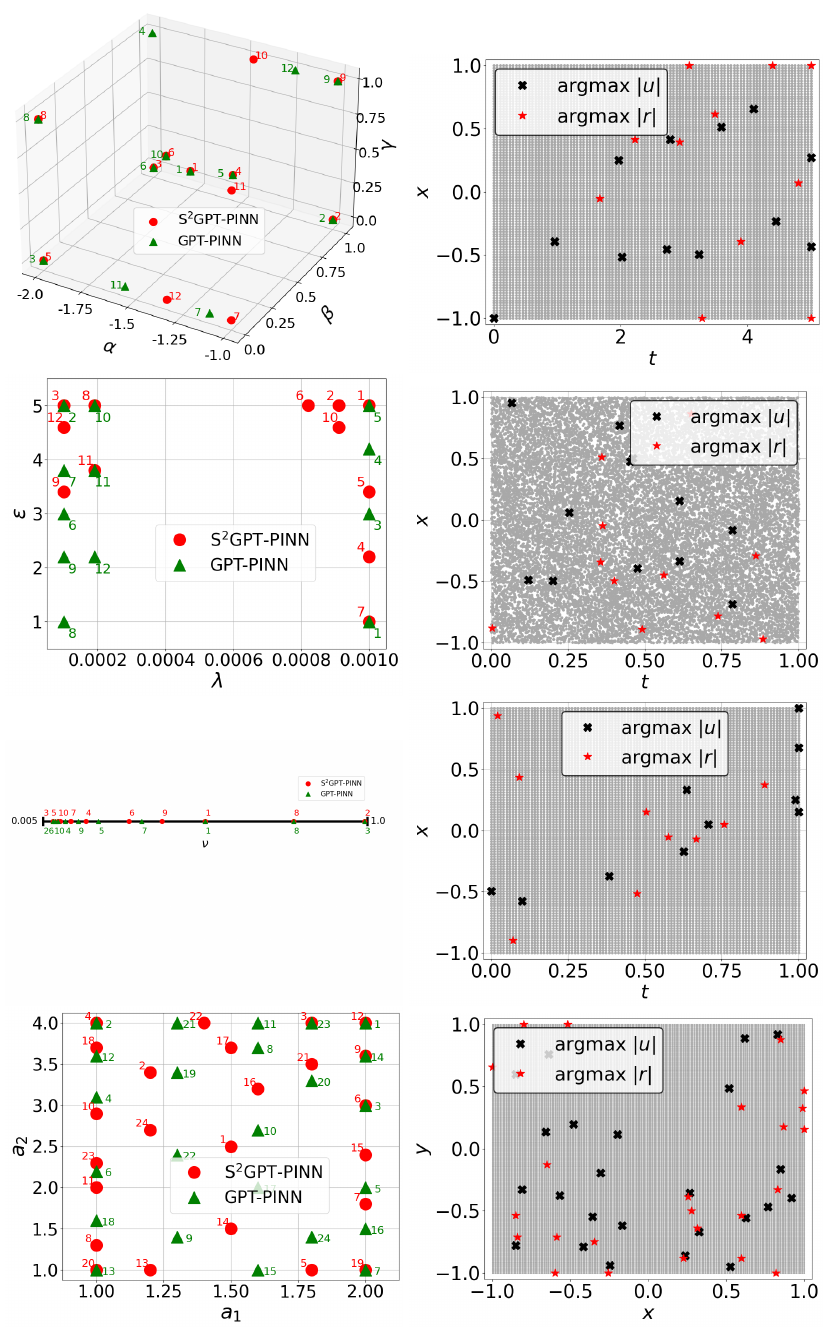}
    \caption{S$^2$GPT-PINN is much sparser than GPT-PINN. Left: Neuron parameters generated by the GPT-PINN (green) and S$^2$GPT-PINN (red). Right: Sparse grid generated by the S$^2$GPT-PINN overlaid with the full grid of GPT-PINN.}
    \label{fig:kgacb-select}
\end{figure}

We display in \Cref{fig:kgacb-select} the parameter values chosen by GPT-PINN and S$^2$GPT-PINN. It is clear that they are unstructured and tend to cluster around the boundary of the parameter domain. 
We also show the collocation grid constructed by S$^2$GPT-PINN. To emphasize how sparse it is in comparison to the full grid adopted by GPT-PINN, we overlay the full grid in the background. The orders of magnitude in sparsification is very noticeable. 

\begin{figure}
    \centering
\includegraphics[width=\linewidth]{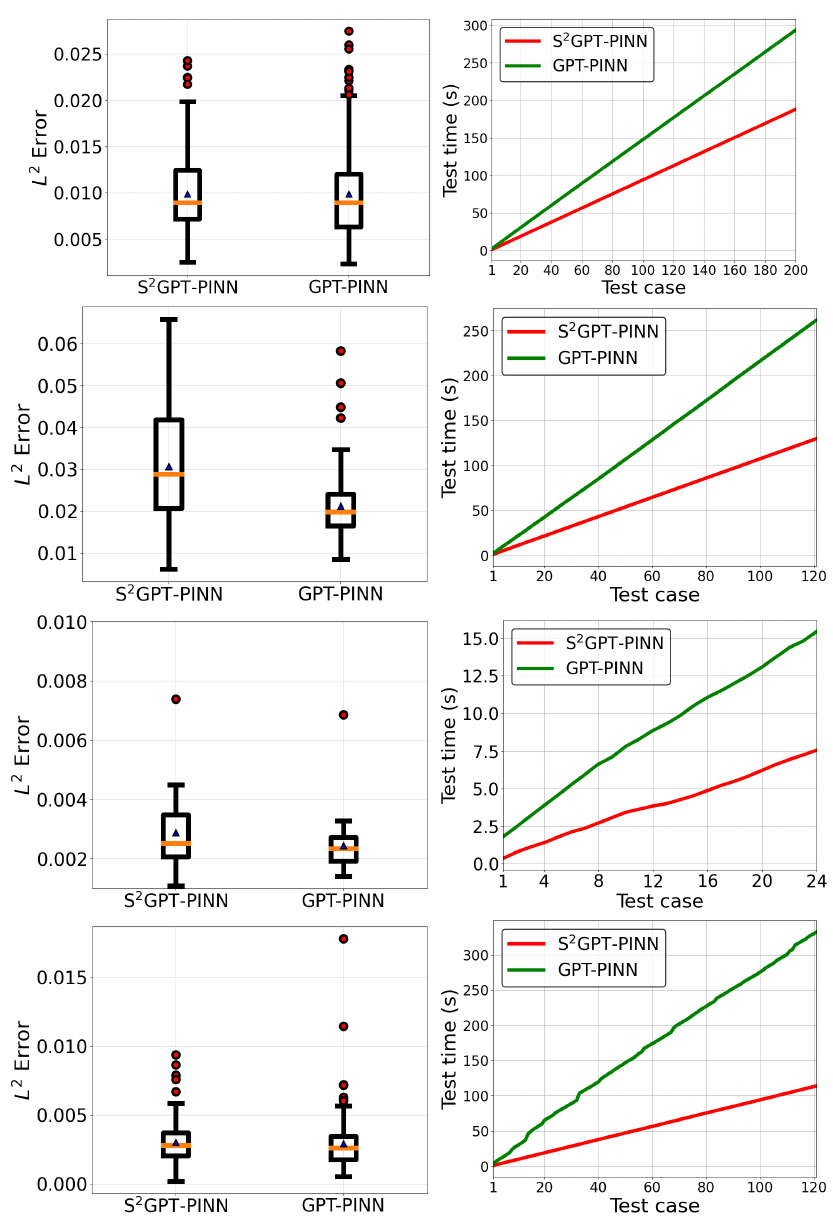}
    \caption{While having comparable accuracy, S$^2$GPT-PINN is much faster than GPT-PINN. Left: Box plot  of L2 errors over all tested parameters. Right: The testing time of GPT-PINN and S$^2$GPT-PINN on a single-core CPU ({focusing on an accurate baseline efficiency comparison under standardized conditions without hardware-specific optimizations}).}
    \label{fig:kgacb-test}
\end{figure}

Finally, to demonstrate that S$^2$GPT-PINN is faster than GPT-PINN in practice while being similarly accurate, we {show} in \Cref{fig:kgacb-test} the cumulative runtime. It is clear that S$^2$GPT-PINN not only outperforms GPT-PINN in offline training efficiency but also is two to three times faster than GPT-PINN thanks to the judicious sparsification strategy.

\section{Conclusion}
\label{sec:conclusion}

S$^2$GPT-PINN, a sparse and small model for solving parametric PDEs, is shown to be as accurate as but faster than its dense version GPT-PINN that was demonstrated to be an effective surrogate of the large model PINN. S$^2$GPT-PINN is tailored to domain-specific tasks and features a compact architecture and minimal computational power. Its main novelty are that it only leverages a small amount of extremely high-quality data, and that it seamlessly integrates two levels of customizations. The first is task-specific activation functions that are transferred from Pre-Trained PINNs while the second is down-sampling by orders of magnitude without losing accuracy. Ongoing works include the development of the nonlinear model reduction version and theoretical analysis. Furthermore, although we have tested our sparsification approach on a variety of problems, its stability when applied to more complex or chaotic systems still requires further investigation.


\begin{thebibliography}{10}

\bibitem{abdin2024phi}
M.~Abdin, J.~Aneja, H.~Awadalla, A.~Awadallah, A.~A. Awan, N.~Bach, A.~Bahree, A.~Bakhtiari, J.~Bao, H.~Behl, et~al.
\newblock Phi-3 technical report: A highly capable language model locally on your phone.
\newblock {\em arXiv:2404.14219}, 2024.

\bibitem{ainsworth2021galerkin}
M.~Ainsworth and J.~Dong.
\newblock Galerkin neural networks: A framework for approximating variational equations with error control.
\newblock {\em SIAM Journal on Scientific Computing}, 43(4):A2474--A2501, 2021.

\bibitem{Barrault_Nguyen_Maday_Patera}
M.~Barrault, N.~C. Nguyen, Y.~Maday, and A.~T. Patera.
\newblock {An ``empirical interpolation'' method: Application to efficient reduced-basis discretization of partial differential equations}.
\newblock {\em C. R. Acad. Sci. Paris, S{\'{e}}rie I}, 339:667--672, 2004.

\bibitem{BinevCohenDahmenDevorePetrovaWojtaszczyk}
P.~Binev, A.~Cohen, W.~Dahmen, R.~Devore, G.~Petrova, and P.~Wojtaszczyk.
\newblock Convergence rates for greedy algorithms in reduced basis methods.
\newblock {\em SIAM Journal on Mathematical Analysis}, 43(3):1457--1472, 2011.

\bibitem{brown2020language}
T.~Brown, B.~Mann, N.~Ryder, M.~Subbiah, J.~D. Kaplan, P.~Dhariwal, A.~Neelakantan, P.~Shyam, G.~Sastry, A.~Askell, et~al.
\newblock Language models are few-shot learners.
\newblock {\em Advances in neural information processing systems}, 33:1877--1901, 2020.

\bibitem{chen2021R2ROC}
Y.~Chen, S.~Gottlieb, L.~Ji, and Y.~Maday.
\newblock {An EIM-degradation free reduced basis method via over collocation and residual hyper reduction-based error estimation}.
\newblock {\em Journal of Computational Physics}, 444:110545, 2021.

\bibitem{chen2024tgpt}
Y.~Chen, Y.~Ji, A.~Narayan, and Z.~Xu.
\newblock {TGPT-PINN: Nonlinear model reduction with transformed GPT-PINNs}.
\newblock {\em Computer Methods in Applied Mechanics and Engineering}, 430:117198, 2024.

\bibitem{chen2024gpt}
Y.~Chen and S.~Koohy.
\newblock {GPT-PINN}: Generative pre-trained physics-informed neural networks toward non-intrusive meta-learning of parametric {PDE}s.
\newblock {\em Finite Elements in Analysis and Design}, 228:104047, 2024.

\bibitem{deng2009imagenet}
J.~Deng, W.~Dong, R.~Socher, L.-J. Li, K.~Li, and L.~Fei-Fei.
\newblock Imagenet: A large-scale hierarchical image database.
\newblock In {\em 2009 IEEE conference on computer vision and pattern recognition}, pages 248--255. Ieee, 2009.

\bibitem{devore2021neural}
R.~DeVore, B.~Hanin, and G.~Petrova.
\newblock Neural network approximation.
\newblock {\em Acta Numerica}, 30:327--444, 2021.

\bibitem{dhariwal2021diffusion}
P.~Dhariwal and A.~Nichol.
\newblock Diffusion models beat gans on image synthesis.
\newblock {\em Advances in neural information processing systems}, 34:8780--8794, 2021.

\bibitem{fresca2021comprehensive}
S.~Fresca, L.~Dede’, and A.~Manzoni.
\newblock A comprehensive deep learning-based approach to reduced order modeling of nonlinear time-dependent parametrized pdes.
\newblock {\em Journal of Scientific Computing}, 87:1--36, 2021.

\bibitem{fresca2022pod}
S.~Fresca and A.~Manzoni.
\newblock {POD-DL-ROM: Enhancing deep learning-based reduced order models for nonlinear parametrized PDEs by proper orthogonal decomposition}.
\newblock {\em Computer Methods in Applied Mechanics and Engineering}, 388:114181, 2022.

\bibitem{gao2023failure}
Z.~Gao, L.~Yan, and T.~Zhou.
\newblock Failure-informed adaptive sampling for {PINNs}.
\newblock {\em SIAM Journal on Scientific Computing}, 45(4):A1971--A1994, 2023.

\bibitem{HanJentzenE2018}
J.~Han, A.~Jentzen, and W.~E.
\newblock Solving high-dimensional partial differential equations using deep learning.
\newblock {\em Proceedings of the National Academy of Sciences}, 115(34):8505--8510, 2018.

\bibitem{he2016deep}
K.~He, X.~Zhang, S.~Ren, and J.~Sun.
\newblock Deep residual learning for image recognition.
\newblock In {\em Proceedings of the IEEE conference on computer vision and pattern recognition}, pages 770--778, 2016.

\bibitem{hesthaven2016certified}
J.~S. Hesthaven, G.~Rozza, B.~Stamm, et~al.
\newblock {\em Certified reduced basis methods for parametrized partial differential equations}, volume 590.
\newblock Springer, 2016.

\bibitem{hesthaven2018nonNNRBM}
J.~S. Hesthaven and S.~Ubbiali.
\newblock Non-intrusive reduced order modeling of nonlinear problems using neural networks.
\newblock {\em Journal of Computational Physics}, 363:55--78, 2018.

\bibitem{hou2023enhancing}
J.~Hou, Y.~Li, and S.~Ying.
\newblock Enhancing pinns for solving pdes via adaptive collocation point movement and adaptive loss weighting.
\newblock {\em Nonlinear Dynamics}, 111(16):15233--15261, 2023.

\bibitem{huang2022meta}
X.~Huang, Z.~Ye, H.~Liu, S.~Ji, Z.~Wang, K.~Yang, Y.~Li, M.~Wang, H.~Chu, F.~Yu, et~al.
\newblock Meta-auto-decoder for solving parametric partial differential equations.
\newblock {\em Advances in Neural Information Processing Systems}, 35:23426--23438, 2022.

\bibitem{ji2025egptpinn}
Y.~Ji, Y.~Chen, and Z.~Xu.
\newblock {EGPT-PINN: Entropy-Enhanced Generative Pre-Trained Physics Informed Neural Networks for Parameterized Nonlinear Conservation Laws}.
\newblock {\em Available at SSRN: \url{https://ssrn.com/abstract=5216939}}, 2025.

\bibitem{lau2024pinnacle}
G.~K.~R. Lau, A.~Hemachandra, S.-K. Ng, and B.~K.~H. Low.
\newblock {PINNACLE}: {PINN} adaptive collocation and experimental points selection.
\newblock In {\em The Twelfth International Conference on Learning Representations}, 2024.

\bibitem{li2021fourier}
Z.~Li, N.~B. Kovachki, K.~Azizzadenesheli, B.~liu, K.~Bhattacharya, A.~Stuart, and A.~Anandkumar.
\newblock Fourier neural operator for parametric partial differential equations.
\newblock In {\em International Conference on Learning Representations}, 2021.

\bibitem{lu2021learning}
L.~Lu, P.~Jin, G.~Pang, Z.~Zhang, and G.~E. Karniadakis.
\newblock Learning nonlinear operators via deeponet based on the universal approximation theorem of operators.
\newblock {\em Nature machine intelligence}, 3(3):218--229, 2021.

\bibitem{lu2024blending}
X.~Lu, Z.~Liu, A.~Liusie, V.~Raina, V.~Mudupalli, Y.~Zhang, and W.~Beauchamp.
\newblock Blending is all you need: Cheaper, better alternative to trillion-parameters llm.
\newblock {\em arXiv:2401.02994}, 2024.

\bibitem{lu2024small}
Z.~Lu, X.~Li, D.~Cai, R.~Yi, F.~Liu, X.~Zhang, N.~D. Lane, and M.~Xu.
\newblock Small language models: Survey, measurements, and insights.
\newblock {\em arXiv:2409.15790}, 2024.

\bibitem{MadayMula2013}
Y.~Maday and O.~Mula.
\newblock {A generalized empirical interpolation method: application of reduced basis techniques to data assimilation}.
\newblock In {\em Analysis and numerics of partial differential equations}, volume~4 of {\em Springer INdAM Ser.}, pages 221--235. Springer, Milan, 2013.

\bibitem{maday2015generalized}
Y.~Maday, O.~Mula, A.~T. Patera, and M.~Yano.
\newblock The generalized empirical interpolation method: stability theory on hilbert spaces with an application to the stokes equation.
\newblock {\em Computer Methods in Applied Mechanics and Engineering}, 287:310--334, 2015.

\bibitem{MCCLENNY2023111722}
L.~D. McClenny and U.~M. Braga-Neto.
\newblock Self-adaptive physics-informed neural networks.
\newblock {\em Journal of Computational Physics}, 474:111722, 2023.

\bibitem{penwardena2021physics}
M.~Penwardena, S.~Zheb, A.~Narayanc, and R.~M. Kirbya.
\newblock {Physics-Informed Neural Networks (PINNs) for Parameterized PDEs: A Metalearning Approach}.
\newblock {\em arXiv:2110.13361}, 2021.

\bibitem{quarteroni_reduced_2016}
A.~Quarteroni, A.~Manzoni, and F.~Negri.
\newblock {\em Reduced {Basis} {Methods} for {Partial} {Differential} {Equations}}.
\newblock Springer International Publishing, 2016.

\bibitem{raissi2019physics}
M.~Raissi, P.~Perdikaris, and G.~E. Karniadakis.
\newblock Physics-informed neural networks: A deep learning framework for solving forward and inverse problems involving nonlinear partial differential equations.
\newblock {\em Journal of Computational Physics}, 378:686--707, 2019.

\bibitem{schick2020s}
T.~Schick and H.~Sch{\"u}tze.
\newblock It's not just size that matters: Small language models are also few-shot learners.
\newblock {\em arXiv:2009.07118}, 2020.

\bibitem{siegel2023greedy}
J.~W. Siegel, Q.~Hong, X.~Jin, W.~Hao, and J.~Xu.
\newblock Greedy training algorithms for neural networks and applications to pdes.
\newblock {\em Journal of Computational Physics}, 484:112084, 2023.

\bibitem{vaswani2017attentionNew}
A.~Vaswani, N.~Shazeer, N.~Parmar, J.~Uszkoreit, L.~Jones, A.~N. Gomez, {\L}.~Kaiser, and I.~Polosukhin.
\newblock Attention is all you need.
\newblock {\em Advances in neural information processing systems}, 30, 2017.

\end{thebibliography}
\bibliographystyle{abbrv}

\end{document}